\documentclass[a4paper, 10pt, conference]{ieeeconf}      

\IEEEoverridecommandlockouts                              

\usepackage[utf8]{inputenc}
\usepackage{dblfloatfix}

\usepackage{graphicx} 
\usepackage{subcaption}
\usepackage{balance}
\usepackage{mathptmx} 
\usepackage{times} 
\usepackage{amsmath} 
\usepackage{amssymb}  
\usepackage{cite}
\usepackage{caption}
\usepackage{xcolor}
\usepackage{multirow}
\usepackage{siunitx}
\usepackage{hyperref}

\usepackage{algorithm}
\usepackage{algpseudocode}
\usepackage{placeins}
\usepackage{booktabs}
\usepackage[super]{nth}
\usepackage[colorinlistoftodos]{todonotes}

\title{\LARGE \bf
Hierarchical Multi-Channel Tip-Over Prediction via Isolated Temporal Models for Planetary Rover Safety
}
\author{
Siddhant Shete$^{1}$, Raúl Domínguez$^{1}$, Frank Kirchner$^{1,2}$%
\thanks{$^{1}$All authors are with the Robotics Innovation Center of the German Research Center for Artificial Intelligence GmbH (DFKI), Bremen, Germany.}%
\thanks{$^{2}$Prof. Frank Kirchner is additionally affiliated with the University of Bremen, Germany.}%
}

\begin{document}

\maketitle
\thispagestyle{empty}
\pagestyle{empty}

\begin{abstract}

Reliable tip-over prevention is critical for planetary rovers operating in unstructured environments where communication delays prevent real-time human intervention. Existing learning-based approaches typically optimize a single prediction objective, limiting their ability to capture both gradual instability evolution and abrupt terrain disturbances. This work presents a hierarchical multi-channel tip-over prediction framework based on isolated temporal learning. A multi-horizon Temporal Convolutional Network (TCN) learns progressive instability patterns from inertial measurements, while a Gated Recurrent Unit (GRU)-based forward dynamics model predicts nominal rover motion and detects unexpected dynamic deviations through prediction residuals. These complementary signals are integrated through decision-level fusion and priority-based arbitration to generate graded safety actions, including early warning and emergency braking. The proposed system is deployed as an Orocos Real-Time Toolkit (RTT)-based safety component within the Rock robotics framework, operating at 100 Hz on a planetary rover software stack. Experiments on Inertial Measurement Unit (IMU) samples collected across heterogeneous terrains demonstrate 97.58\% event prediction coverage with 1.82s median lead time, outperforming individual prediction channels. The results validate that isolated temporal models combined through hierarchical decision fusion provide a robust solution for real-time rover safety in uncertain terrain.

\end{abstract}


\section{INTRODUCTION}

\begin{figure}[h]
    \centering
    
    \begin{subfigure}{0.43\columnwidth}
        \centering
        \includegraphics[width=\linewidth]{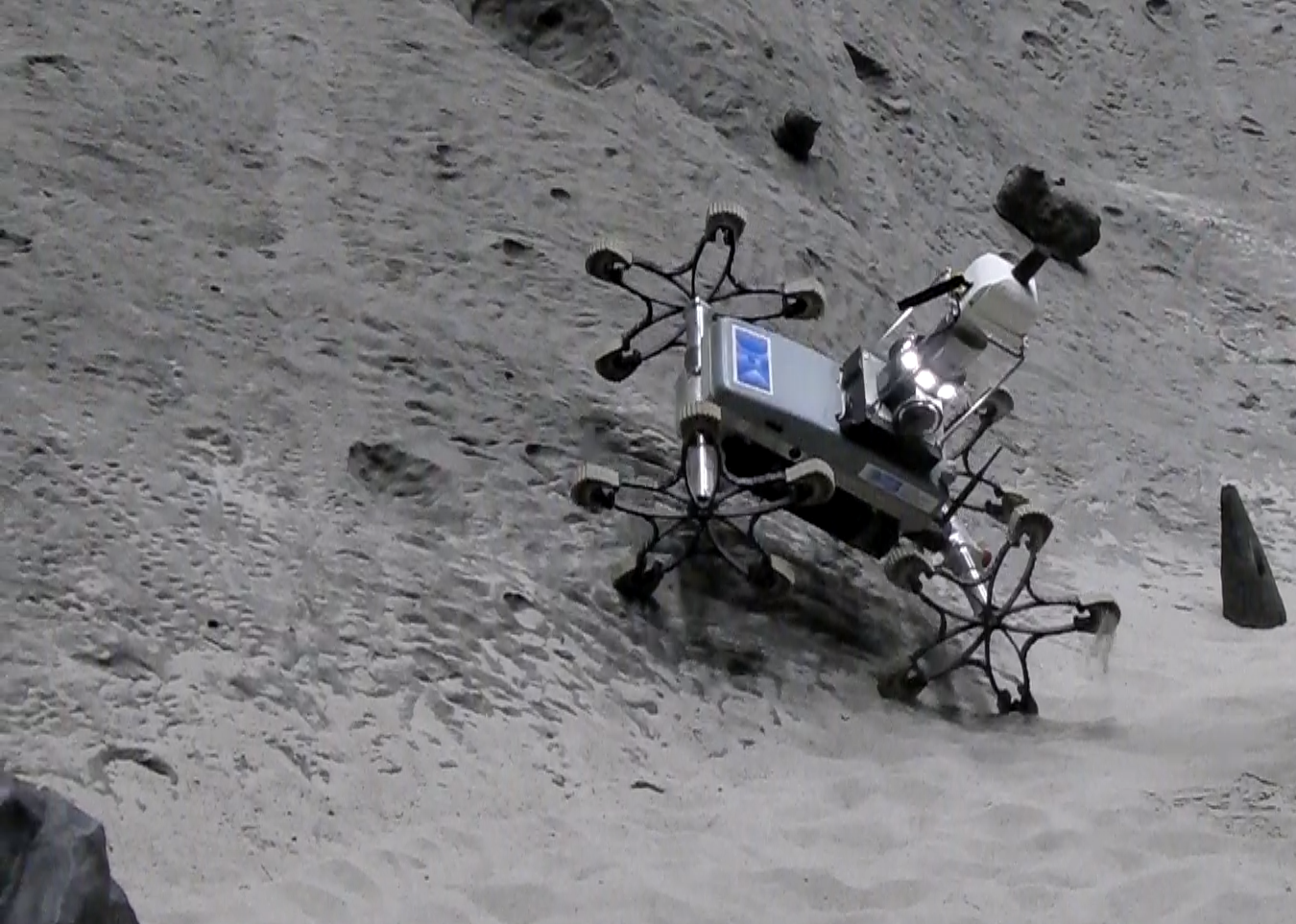}
        \caption{Before tip-over}
        \label{fig:rover_before}
    \end{subfigure}
    \hfill
    \begin{subfigure}{0.48\columnwidth}
        \centering
        \includegraphics[width=\linewidth]{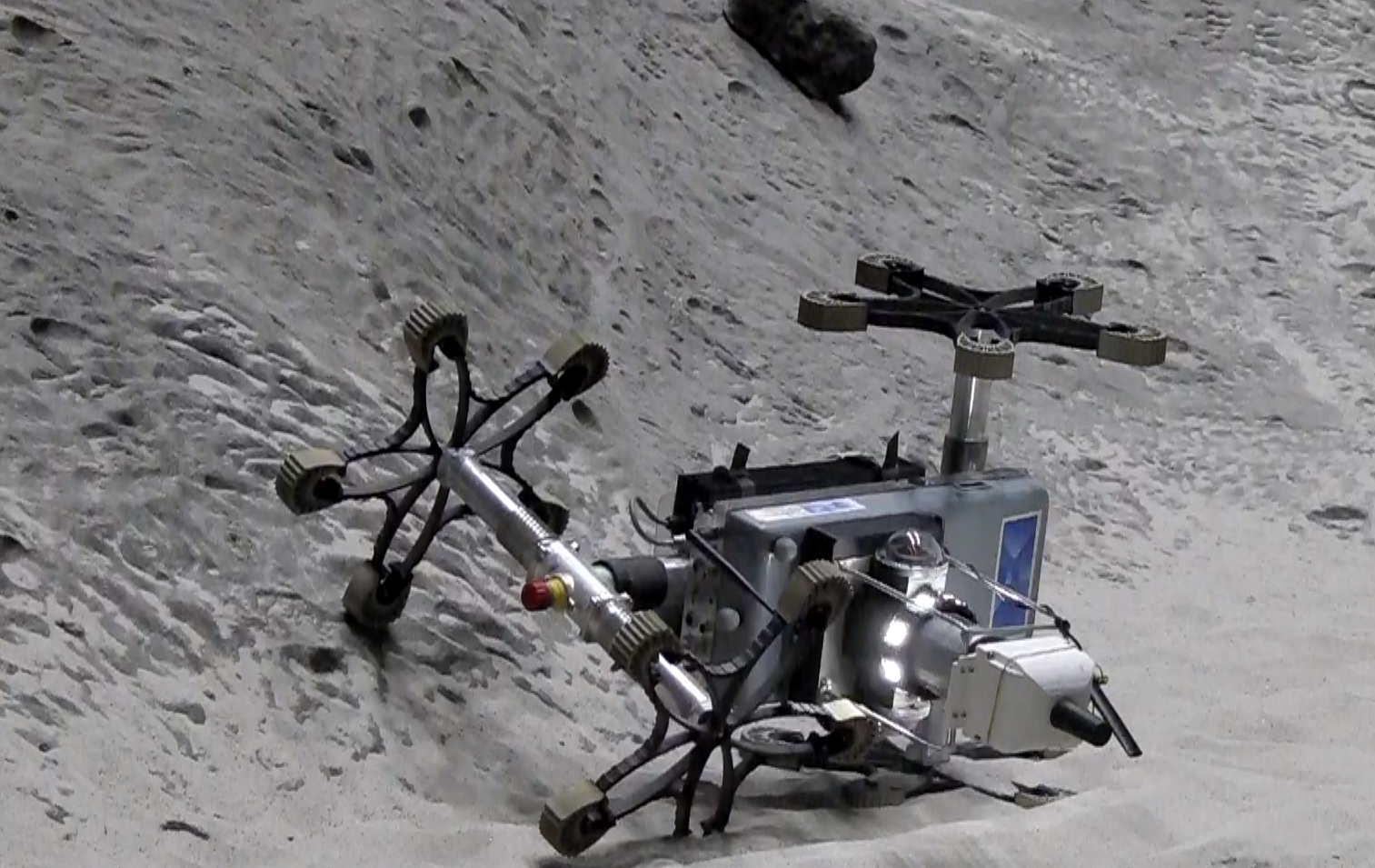}
        \caption{After tip-over}
        \label{fig:rover_after}
    \end{subfigure}

    \caption{Asguardv4 rover during a tip-over event while climbing an inclined slope.}
    \label{fig:asguard}
\end{figure}

Planetary rovers must operate autonomously in highly unstructured terrain while maintaining stability, as loss of balance can lead to irreversible mission failure. In space missions, one-way communication delays between Earth and Mars vary from approximately 4 to 24 minutes depending on the relative planetary positions, preventing real-time human intervention, requiring onboard systems capable of autonomous safety assessment and rapid hazard response \cite{Mars2020}. These constraints make predictive tip-over prevention a critical component of planetary rover autonomy. In practice, rover navigation systems rely on pre-planned trajectories combined with reactive hazard avoidance. However, unstructured environments introduce unforeseen disturbances, such as sudden rock impacts, underlying voids, and distinct high-slip areas that can induce rapid destabilization \cite{Reviewonspacerobotics}, as demonstrated in Fig. \ref{fig:asguard}. Although rovers are designed with considerable static stability margins, dynamic instability caused by terrain interactions (e.g., wheel slip) can develop before conventional stability margins indicate imminent failure. Therefore, autonomous rover systems require predictive models capable of reasoning about the temporal evolution of instability rather than only evaluating instantaneous stability states.

The key challenge is that rover instability does not evolve through a single temporal pattern. Some failure scenarios arise from gradually changing rover–terrain interactions that progressively drive the system toward instability, whereas others are triggered by sudden disturbances such as wheel slip, unexpected impacts, or loss of ground support \cite{Reviewonspacerobotics,Rose}, leaving little opportunity for intervention. These two instability mechanisms require different temporal reasoning capabilities. Predictive tip-over classification must identify discriminative patterns associated with evolving hazards, while forward dynamics prediction must accurately represent nominal rover motion and detect deviations from expected behavior. Consequently, forcing both objectives into a shared representation may reduce the ability of individual models to specialize for their respective safety tasks. Motivated by this observation, we learn these objectives independently through isolated temporal models, allowing each branch to specialize in its safety-critical objective while avoiding unnecessary interference between prediction tasks. To address these limitations, we propose a hierarchical multi-channel tip-over prediction framework that employs isolated temporal representations within a unified learning architecture. A TCN learns multi-horizon instability evolution, while a GRU forward model captures deviations from nominal rover dynamics. Their outputs are combined only at the decision level through hierarchical fusion to generate graded safety actions.

The main contributions of this work are (1) an isolated dual-temporal architecture separating predictive instability assessment from nominal dynamics consistency estimation;
(2) a hierarchical decision fusion mechanism combining predictive confidence and dynamics residuals for graded safety actions; and
(3) a real-time onboard implementation within the Rock robotics framework\cite{pub11033}, evaluating real-time tip-over prediction and safety arbitration on a planetary rover platform operating at 100 Hz.

Although evaluated on a planetary rover software stack, the proposed architecture provides a general strategy for integrating heterogeneous predictive safety signals in autonomous robotic systems where timely and reliable intervention is required. The separation of predictive temporal learning and dynamics-consistency estimation through isolated models provides a general design principle for combining complementary safety signals while avoiding the optimization conflicts associated with shared multi-task representations.


\section{RELATED WORK}
Classical rover stability assessment relies on analytical criteria derived from robot-terrain interaction models. The Zero-Moment Point (ZMP) criterion evaluates dynamic stability by constraining the resultant ground reaction moment within the support polygon \cite{ZMP}, while static stability margins provide computationally efficient geometric stability guarantees under quasi-static assumptions \cite{MCGHEE1968331}. The Moment-Height Stability (MHS) index further incorporates inertial effects for wheeled robotic systems while maintaining low computational complexity \cite{Fuentes}. However, these analytical approaches depend on accurate terrain-contact assumptions and simplified interaction models, which can become unreliable under wheel slip, loose soil, unexpected impacts, and highly deformable planetary terrain\cite{Mars2020}. Recent rover mobility studies have demonstrated that stability thresholds vary across terrain conditions, limiting the generalization of fixed analytical criteria for complex environments \cite{shete}. Meanwhile, planetary rover autonomy has advanced toward integrated perception, terrain reasoning, and adaptive navigation under communication constraints and environmental uncertainty \cite{Reviewonspacerobotics, Fong}. However, these approaches primarily address traversability and path planning rather than predictive vehicle stability estimation.

Recent advances in onboard computation and inertial sensing have enabled learning-based approaches for predicting future robotic states directly from temporal sensor observations. Recurrent architectures such as Long Short-Term Memory (LSTM) and GRU can capture nonlinear temporal dependencies in robot dynamics, while TCNs provide efficient long-range sequence modeling with low inference latency \cite{Iwamoto,Shaojie}. Recent learning-based approaches for mobile robots and off-road vehicles demonstrate improved early hazard anticipation compared with threshold-based monitoring \cite{Rose,shete}. Learning-based mobility estimation has improved terrain traversability, wheel slip prediction, and vehicle state estimation in off-road environments \cite{Manderson, Wellhausen_2020}. Most existing learning-based approaches optimize a single predictive objective, such as instability classification or dynamics forecasting. Consequently, they capture only one aspect of rover instability, limiting their ability to jointly model gradual instability evolution and abrupt terrain-induced disturbances.

Prediction-error-based anomaly detection learns nominal dynamics and identifies deviations through future prediction residuals. LSTM encoder-decoder models have been widely applied for multivariate sensor anomaly detection \cite{malhotra2016lstmbasedencoderdecodermultisensoranomaly}, while probabilistic approaches improve robustness by modeling nominal behavior uncertainty \cite{Pang_2021, Kang_2025}. Such methods reduce dependence on failure annotations \cite{rauch2013comparison}. However, prediction residuals are inherently ambiguous: large deviations may indicate imminent instability or simply unfamiliar yet recoverable rover–terrain interactions. This ambiguity limits their reliability as standalone safety triggers in highly variable planetary environments. Safety-critical robotics further employs runtime monitoring and uncertainty-aware decision frameworks for robust operation under uncertain environments \cite{fisac2018,Althoff,kendall2017}; however, these approaches generally supervise detected hazards rather than predicting instability evolution.

Collectively, existing approaches provide complementary capabilities, including analytical stability estimation, temporal prediction, anomaly detection, and safety monitoring. However, these capabilities are typically developed in isolation and are rarely integrated into a unified prediction architecture for autonomous rover safety. While multi-task learning could combine multiple objectives within a shared model, shared representations may introduce optimization conflicts between discriminative classification and predictive dynamics modeling \cite{yu2020gradientsurgerymultitasklearning}. The proposed framework addresses this limitation by learning independent temporal representations and combining their outputs through hierarchical decision-level arbitration, preserving task-specific specialization while maintaining an interpretable and computationally efficient safety framework.

\section{System and data acquisition}
The proposed framework targets deployment on the LuNiS\footnote{\url{https://robotik.dfki-bremen.de/de/forschung/robotersysteme/lunis}} planetary rover, a terrestrial technology demonstrator developed to validate autonomous capabilities and subsystem integration for future lunar micro-rover missions. Since collecting large numbers of real tip-over events on LuNiS risks hardware damage, training and evaluation data were collected using the Asguardv4\footnote{\url{https://robotik.dfki-bremen.de/en/research/robot-systems/asguard-iv}} experimental rover. Asguard was repeatedly driven over slopes, rocks, loose sand, and crater-like terrain at DFKI premises, producing nominal, recovery, and tip-over behaviors across diverse terrain conditions. The final dataset contains 757,316 synchronized IMU samples from multiple experimental sessions, including nominal and hazardous rover dynamics. The data were partitioned in temporal order as training, validation, test, and training blocks, with the two training blocks combined for model development. Sliding windows were generated separately within each partition, and boundary-spanning windows were discarded. The split is therefore time-block based, not session- or terrain-run based. The dataset is publicly available to support reproducibility \cite{shete_2026_21258126}.


\section{methodology}
\begin{figure*}[h]
    \centering
    \includegraphics[width=0.88\linewidth]{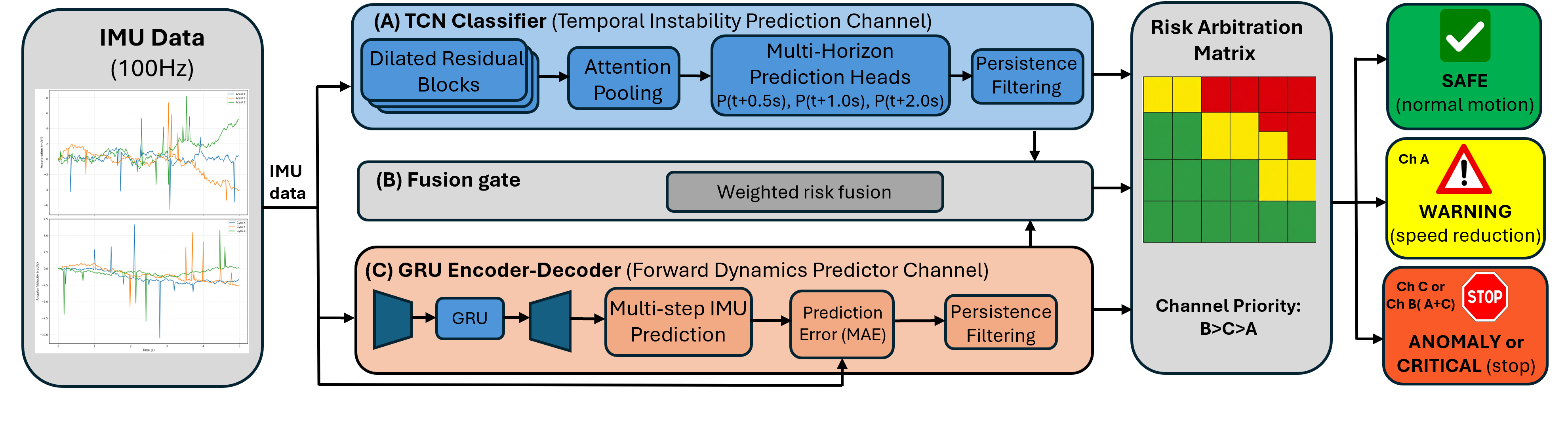}
    \caption{Overview of a multi-channel tip-over prediction framework with TCN, GRU, and fusion-based priority decision for safety actions.}
    \label{fig:arch}
\end{figure*}

Figure~\ref{fig:arch} presents the proposed framework. Two parallel temporal branches process the same IMU sequence: a TCN performs multi-horizon tip-over prediction, whereas a GRU models nominal rover dynamics through forward prediction. Their outputs are combined only at the decision level within a hierarchical three-channel architecture to generate graded safety actions.

\subsection{Preprocessing and Input Representation}

The system processes six-dimensional IMU measurements comprising three-axis linear acceleration and angular velocity sampled at 100 Hz, matching the rover control frequency for continuous onboard safety assessment.  A sliding window of 50 IMU samples forms the temporal input sequence, balancing temporal context, inference latency, and computational efficiency. Although the input window is limited to reduce prediction delay, the dilated temporal convolutions within the TCN expand the effective receptive field, allowing the model to capture both short-term inertial variations and progressive instability patterns.

The same input sequence is provided to two independently trained temporal models with different objectives. Channel A learns discriminative representations for future tip-over prediction using multi-horizon supervision, whereas Channel C learns nominal rover dynamics for prediction-error-based anomaly detection.  To ensure that the GRU learns normal system behavior rather than already unstable motion, its training sequences are extracted with a pre-event gap before each tip-over occurrence. During deployment, both models process streaming IMU measurements in parallel, and their outputs are fused and combined later by the hierarchical decision module.

\subsection{Channel A: Temporal Instability Prediction}
Channel A predicts future tip-over probability by learning temporal instability representations from IMU sequences. It employs a causal TCN with dilated convolutions to efficiently capture long-range temporal dependencies while maintaining deterministic inference latency suitable for real-time deployment. The TCN jointly predicts tip-over probabilities at 0.5s, 1s, and 2s during training, enabling multi-horizon supervision. During deployment, only the 0.5s head is retained for safety decisions because it provides the highest predictive reliability while benefiting from the richer temporal representations learned through auxiliary horizons. The encoder consists of residual causal dilated convolution blocks followed by temporal pooling and lightweight prediction heads. The network produces probabilistic estimates:
$P_{\text{tip}}^{0.5s}(t), P_{\text{tip}}^{1s}(t), P_{\text{tip}}^{2s}(t)$. 
An auxiliary time-to-event regression objective is used during training to provide additional supervision but is not used during inference.

\subsection{Channel C: Forward Dynamics Prediction}
Channel C is a sequence-to-sequence GRU encoder-decoder trained exclusively on nominal rover motion to predict future IMU dynamics from the same input window used by Channel A. During online operation, the model processes overlapping sliding IMU windows at every control cycle, generating a future IMU sequence prediction. Although the complete future sequence is available after each inference, only the first predicted future sample is used to compute the anomaly score. This choice exploits the most reliable short-horizon prediction while avoiding the larger uncertainty associated with later prediction samples, which increases the residual errors during nominal operation and leads to unnecessary false positives. The mean absolute error (MAE) between the first predicted IMU sample and the corresponding measurement forms the prediction residual, yielding a continuously updated anomaly score at 100 Hz. Residuals exceeding a threshold derived from nominal prediction-error statistics activate the anomaly channel, providing rapid terrain-independent detection of unexpected dynamic inconsistencies without requiring explicit terrain models or predefined stability thresholds.



\subsection{Channel B: Decision-Level Fusion}

Neither Channel A nor Channel C alone provides sufficient reliability for safety-critical rover intervention. Channel A predicts gradually evolving instability from learned temporal patterns, providing reliable early warnings but potentially degrading when rover dynamics differ from the training distribution. In contrast, Channel C rapidly detects unexpected motion through elevated prediction residuals, although large residuals may also arise from moderate terrain variations. Consequently, relying on either channel independently may result in false positives or missed hazardous events.

To exploit their complementary strengths while preserving task-specific feature representations, the proposed framework performs fusion exclusively at the decision level. Let $p_t$ denote the tip-over probability estimated by Channel A, and let $\tilde{e}_t$ denote the normalized GRU prediction residual score from Channel C. The fused risk score is computed as:

\[
R_t=\alpha p_t+(1-\alpha)\tilde{e}_t,
\]

where $0 \leq \alpha \leq 1$ controls the contribution of learned instability prediction and dynamics-consistency evidence. In this work, $\alpha=0.85$ was determined on the validation set to emphasize the early prediction capability of Channel A while preserving complementary anomaly information from Channel C. This formulation intentionally avoids additional learned fusion layers, maintaining an interpretable safety decision mechanism while preventing additional optimization dependencies. The fused risk score is filtered using persistence logic before priority-based arbitration generates graded safety actions. Channel A issues early warnings, whereas Channel B or Channel C triggers emergency braking. A lockout interval suppresses repeated emergency commands.

\subsection{System Integration and Real-Time Deployment}
\begin{figure}[h]
    \centering
    \includegraphics[width=\columnwidth]{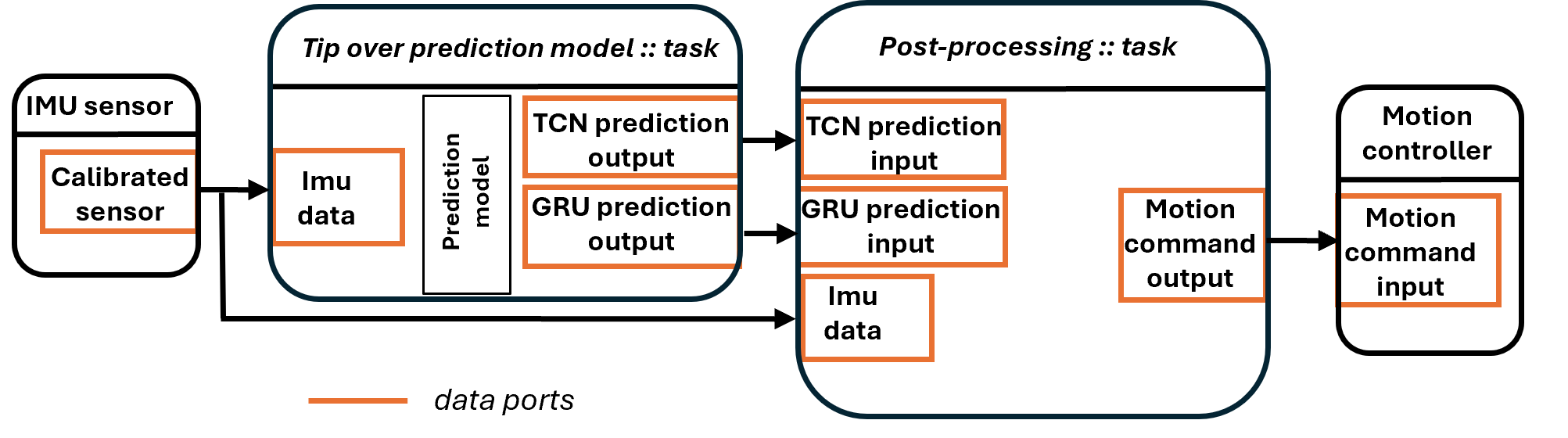}
    \caption{Rock framework illustrating integration of IMU data acquisition, prediction, post-processing, and safety arbitration tasks for real-time rover motion control.}
    \label{fig:rock}
\end{figure}

The proposed framework is implemented within the Robot Construction Kit (Rock)\footnote{\url{https://www.rock-robotics.org/documentation/index.html}} middleware developed at the DFKI Robotics Innovation Center and deployed as an onboard safety component operating at 100 Hz, as shown in Fig. \ref{fig:rock}. Rock was selected because its component-based architecture and integration with the Orocos Real-Time Toolkit (RTT) provide modularity and deterministic execution, enabling reliable integration of sensor streams, neural inference modules, and rover control interfaces. Each functional block shown in Fig. \ref{fig:rock}, including the prediction and safety arbitration modules, is implemented as an independent Orocos RTT task component within Rock, enabling modular scheduling and communication of IMU measurements, prediction outputs, and safety commands through typed data ports. The inference pipeline operates continuously using sliding-window IMU inputs. Sensor data acquisition, neural inference, and decision outputs are synchronized within the execution loop to preserve temporal consistency between the prediction channels and the safety arbitration module. Safety decisions and intermediate prediction signals are logged for post-mission analysis and evaluation. Following emergency activation, a software timeout maintains the braking state for a fixed duration, preventing oscillatory control behavior caused by transient terrain disturbances. The emergency braking command is transmitted to the rover controller after safety arbitration, and the rover requires about 0.5s to complete the commanded stop due to actuator and vehicle dynamics. The prediction lead time is measured from the first system alert to tip-over onset. The nominal braking response of approximately 0.5s is the time from issuing the emergency-brake command to complete standstill. Thus, a median lead time of 1.82s leaves an estimated 1.32 s beyond the nominal stopping response for preventive action.

The complete pipeline, including TCN inference, GRU inference, prediction-residual computation, decision-level fusion, and safety arbitration, features an average execution time of 1.38ms per cycle, making it suitable for real-time onboard deployment.

\section{Experimental Results}
We evaluate the proposed framework on the Asguardv4 rover using sequence-level classification metrics and event-level prediction analysis. The evaluation investigates: (i) the effect of prediction horizon on temporal instability forecasting, (ii) the contribution of individual decision channels and their fusion, and (iii) the trade-off between prediction reliability and available reaction time.

\subsection{Multi-Horizon Predictive Capability}
\begin{figure}[h]
    \centering
    \includegraphics[width=0.5\columnwidth]{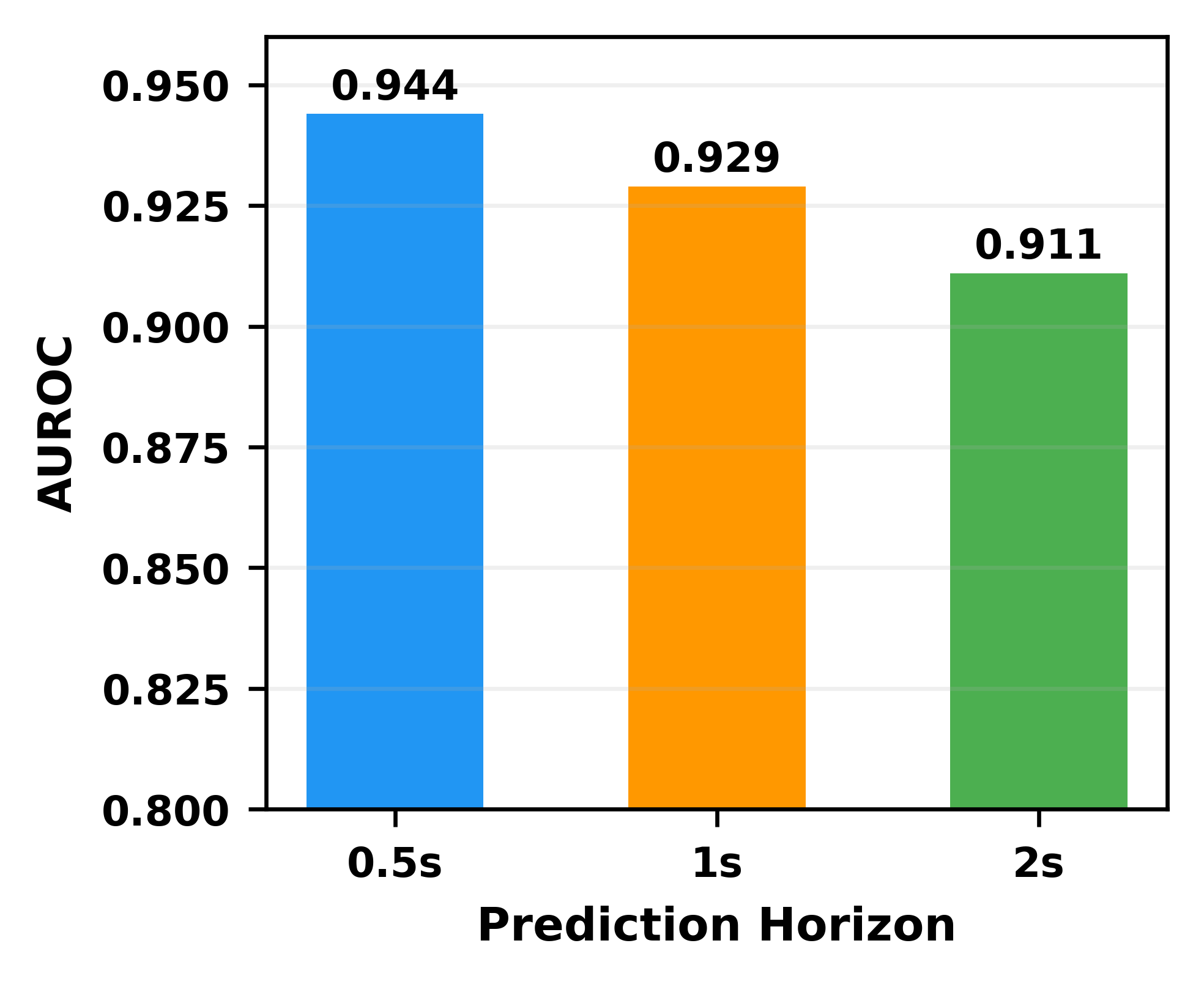}
    \caption{AUROC of the TCN classifier for 0.5s, 1s, and 2s prediction horizons.}
    \label{fig:auroc_horizon}
\end{figure}

To evaluate future instability prediction capability, the TCN classifier is assessed at three forecasting horizons (0.5s, 1s, and 2s ahead). As shown in Fig.~\ref{fig:auroc_horizon}, the prediction performance gradually decreases as the forecasting horizon increases, with the highest performance obtained at the shortest horizon (0.5s). This behavior is expected because uncertainty accumulates when predicting future rover dynamics further ahead. Nevertheless, the model maintains strong discriminative capability even at the 2s horizon (AUROC = 0.911), demonstrating that the learned temporal representation captures both immediate instability precursors and longer-term degradation trends. These results validate the effectiveness of multi-horizon supervision for learning temporally consistent instability features.

\begin{table}[h]
\centering
\caption{Sequence-level and event-level performance comparison of individual prediction channels and the final hierarchical safety output}
\label{tab:results}
\footnotesize
\setlength{\tabcolsep}{4pt}
\begin{tabular}{lccc|ccccc}
\hline
&
\multicolumn{3}{c|}{\textbf{Sequence-wise}} &
\multicolumn{5}{c}{\textbf{Event-wise}} \\
&
\multicolumn{3}{c|}{\textbf{Performance}} &
\multicolumn{5}{c}{\textbf{Performance}} \\
\hline
\textbf{Method} &
\textbf{P} &
\textbf{R} &
\textbf{F1} &
\textbf{EP} &
\textbf{NE} &
\textbf{PE} &
\textbf{Pred.} &
\textbf{MLT(s)} \\
\hline
Ch A(TCN)    & 0.843 & 0.825 & 0.834 & 102 & 8  & 7 & 110/124 & 1.88 \\
Ch B(Fusion) & 0.907 & 0.885 & 0.892 & 99  & 9  & 8 & 108/124 & 1.82 \\
Ch C(GRU)    & 0.639 & 0.795 & 0.703 & 57  & 24 & 8 & 81/124  & 0.74 \\
\hline
\textbf{Hierarchical}
& \textbf{0.934} & \textbf{0.913} & \textbf{0.921}
& \textbf{114} & \textbf{7} & \textbf{0}
& \textbf{121/124} & \textbf{1.82} \\
\textbf{Decision}
& & & & & & & &\\
\hline
\end{tabular}
\vspace{2pt}
\begin{flushleft}
\footnotesize
P=Precision, R=Recall, Pred.=Predicted events, MLT=Median Lead Time (seconds), EP=Early Prediction, NE=Near Event, PE=Post-Event
\end{flushleft}
\end{table}

\begin{figure*}[!b]
    \centering
    \includegraphics[width=0.88\textwidth]{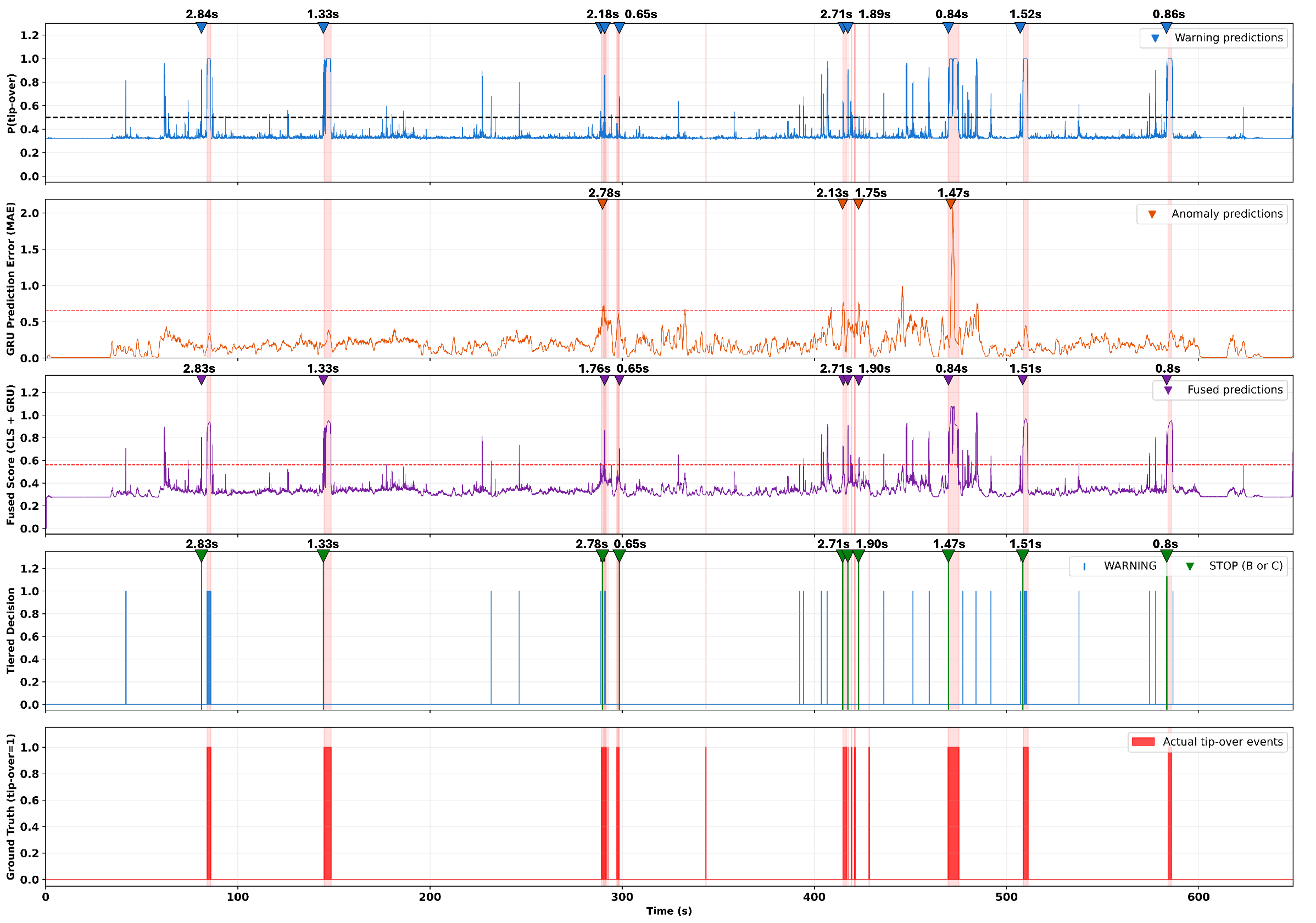}
    \caption{Illustration of the proposed hierarchical multi-channel prediction framework. From top to bottom: Channel A, Channel C, Channel B (fused risk score), tiered safety decisions (early warning and emergency braking), and ground-truth tip-over events. Event-wise prediction lead times are indicated by markers. The initial non-zero Channel A output (\(\sim 0.3\)) is the classifier probability for the first available IMU input window; it is not itself a warning, which is triggered only when the configured threshold and persistence criterion are satisfied.}
    \label{fig:event_analysis}
\end{figure*}

Table~\ref{tab:results} compares the sequence-level classification performance and event-level prediction behavior of the individual channels and the final hierarchical safety output. The results demonstrate the complementary roles of the three prediction pathways within the proposed architecture. Channel A achieves strong sequence-level performance (F1=0.834) with the highest standalone lead time (1.88s) among individual channels, demonstrating its effectiveness for early hazard anticipation.

Channel C achieves lower standalone performance (F1=0.703) and a shorter lead time (0.74s), reflecting its sensitivity to short-duration instability signatures rather than early prediction. Despite its lower standalone performance, Channel C supplies short-horizon anomaly evidence; its shorter lead time reflects its role in detecting sudden dynamic inconsistencies rather than forecasting long-term instability. Channel B combines the probabilistic instability prediction from Channel A with the dynamics-consistency information from Channel C through weighted fusion. The improvement originates from the complementary failure modes of the two temporal models. Channel A learns discriminative patterns associated with progressive instability evolution but may degrade when rover dynamics differ from previously observed instability trajectories. Channel B increases the F1 score to 0.892 while maintaining a lead time of 1.82s, demonstrating that combining independent prediction signals improves reliability compared with either individual channel.

The final hierarchical output further improves performance over Channel B alone, achieving F1=0.921 and detecting 121/124 tip-over events. This improvement results from allowing Channel C to independently trigger intervention during rapid dynamic deviations, while Channel B provides high-confidence decisions when both prediction sources agree. The arbitration strategy therefore improves event coverage without sacrificing the robustness of fused decisions.

The evaluation uses held-out temporal blocks of the dataset and demonstrates performance across the recorded terrain conditions and rover platform, but does not establish generalization to unseen terrains or other rover morphologies. Compared with our previous single-channel GRU-based prediction framework~\cite{shete}, the proposed architecture separates long-horizon instability prediction from short-horizon dynamics inconsistency estimation and combines both through hierarchical decision fusion, reducing false positives while improving event-level coverage and maintaining sufficient reaction time for preventive rover control.


\subsection{Temporal Behavior of the Hierarchical Multi-Channel Predictor}

Figure~\ref{fig:event_analysis} illustrates the temporal interaction between the proposed prediction channels and the final safety arbitration mechanism during a representative tip-over event. The visualization includes the TCN instability confidence, GRU prediction residual, fused risk estimation, final safety decisions, and corresponding ground-truth tip-over interval.

Channel A generates an increase in instability probability before the onset of the ground-truth event, demonstrating its capability to identify gradual degradation trends from inertial dynamics. The blue markers indicate the initial activation of the early-warning mechanism, while the associated annotations show the available lead time before tip-over occurrence. These early predictions provide additional reaction margin for preventive actions, such as reducing rover velocity or modifying the planned trajectory. Channel C responds to rapid deviations between predicted and observed rover dynamics through the GRU prediction residual. The orange markers indicate anomaly-based activations and their corresponding reaction times. Compared with Channel A, these activations occur closer to the instability boundary, reflecting the role of Channel C as a rapid-response mechanism for abrupt terrain-induced disturbances. The fusion channel combines the independent evidence provided by Channels A and C. As shown by the purple markers, the fused decisions exhibit more stable activation behavior than individual signals while maintaining sufficient reaction time for safety intervention. 

The final hierarchical decision layer converts these prediction signals into graded safety actions. Early-warning states generated by Channel A maximize available reaction time, whereas emergency commands triggered by Channel B or Channel C provide reliable emergency braking when instability becomes sufficiently evident. The temporal ordering between warning and emergency decisions demonstrates that the proposed framework performs proactive safety intervention rather than reactive post-event detection.

\subsection{Trade-off between Accuracy and Responsiveness}

\begin{figure}[h]
    \centering
    \includegraphics[width=0.65\linewidth]{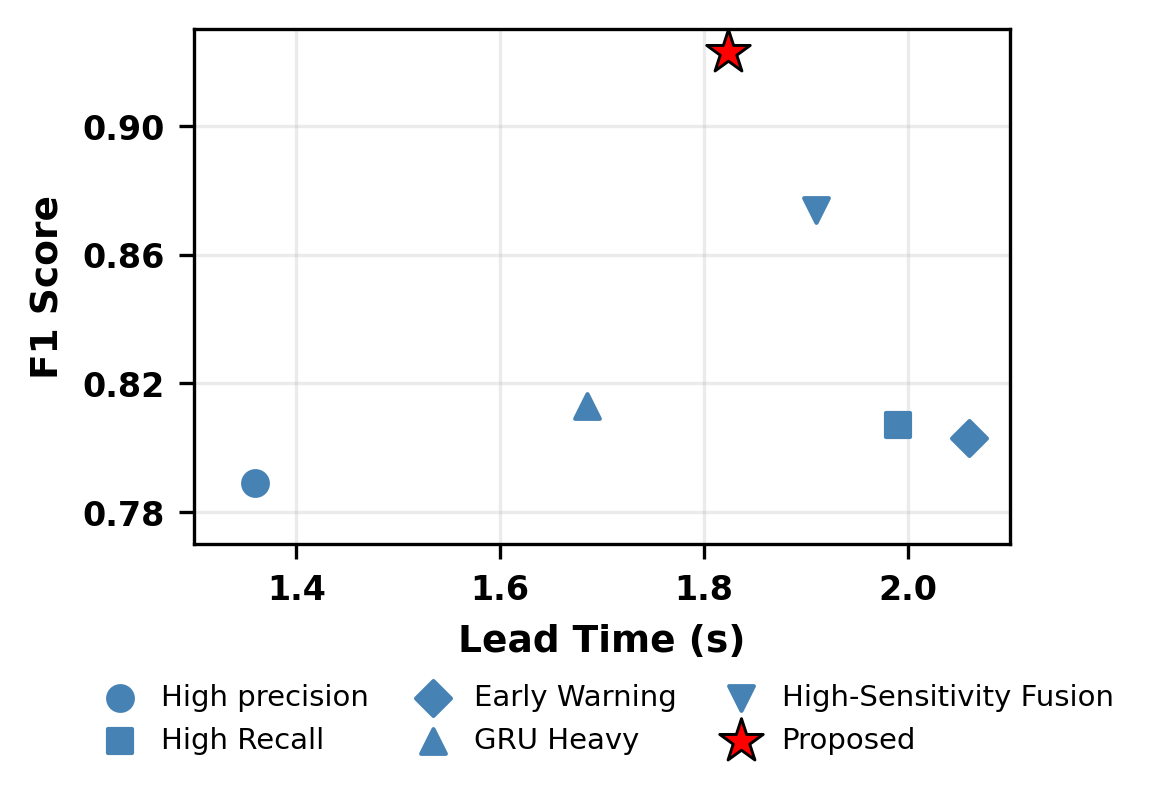}
    \caption{Trade-off between median lead time and F1 for different operating configurations}
    \label{fig:pareto}
\end{figure}

While high prediction accuracy is desirable, autonomous rover safety also requires sufficient reaction time for preventive control actions. The proposed framework is therefore evaluated under multiple operating configurations representing conservative prediction, early warning, high recall, and sensitivity-oriented fusion. Figure~\ref{fig:pareto} summarizes the relationship between sequence-level F1 score and median event lead time. The results demonstrate the inherent trade-off between responsiveness and reliability. Configurations optimized for earlier intervention provide longer lead times but increase false activations, whereas conservative configurations improve reliability with reduced reaction margin. This trade-off highlights the challenge of balancing mobility preservation and safety intervention in dynamic planetary environments.

Among the evaluated configurations, the proposed hierarchical strategy provides the most favorable operating point, achieving an F1 score of 0.921 with a median lead time of 1.82s. This demonstrates that the proposed architecture maintains reliable prediction performance while preserving sufficient temporal margin for emergency rover actions. After accounting for the approximately 0.5s rover braking response, the proposed system retains more than 1.3s of effective intervention margin before tip-over. The resulting operating point lies close to the Pareto-optimal region, indicating that further improvements in reaction time would require a significant reduction in prediction reliability. This balanced behavior is critical for autonomous rover operation, where excessive conservatism may limit mobility while delayed intervention can result in irreversible instability.



\section{Conclusion and future work}
We presented a hierarchical multi-channel tip-over prediction framework that combines isolated temporal learning with decision-level fusion for real-time rover safety. Independent TCN and GRU models capture progressive instability evolution and short-term dynamics inconsistency, respectively, while hierarchical arbitration converts complementary predictions into graded safety actions. The experimental evaluation demonstrates that the proposed framework provides more reliable tip-over prediction on the evaluated dataset compared with individual prediction channels while maintaining low computational latency suitable for onboard real-time operation. The final hierarchical output achieved an F1 score of 0.921, providing 97.58\% event coverage by detecting 121 out of 124 tip-over events with a median lead time of 1.82s before instability. Considering the approximately 0.5s rover braking response time, more than 1.3s of effective intervention margin remains available from the measured 1.82 s lead time, leaving an estimated 1.32 s beyond the nominal braking response. The improvement over single-channel prediction demonstrates the advantage of combining complementary temporal representations: the classifier provides semantic evidence of evolving instability, whereas the dynamics model provides rapid sensitivity to unexpected motion deviations. Furthermore, the Pareto analysis confirms that the proposed approach achieves a favorable balance between prediction reliability and responsiveness, reducing unnecessary interventions while preserving adequate safety margins.

The proposed framework was implemented as an Orocos RTT task component within the Rock robotics framework, operating at 100 Hz on a planetary rover platform, achieving millisecond-scale inference latency and demonstrating its suitability for onboard autonomous operation. However, false activations may still occur during highly dynamic yet recoverable terrain interactions, where the boundary between safe transient maneuvers and genuinely hazardous instability becomes difficult to distinguish from inertial measurements alone. Although decision-level fusion significantly reduces such ambiguities compared with single-model approaches, generalization to unseen terrains and other rover morphologies has not yet been evaluated, given the limited availability of real planetary mobility datasets.

Future work will focus on extending the proposed framework toward a more comprehensive risk-aware rover autonomy architecture. Explicit time-to-event estimation will be investigated to provide continuous instability forecasts beyond discrete safety thresholds, enabling adaptive intervention strategies. Additionally, integration with visual terrain perception and uncertainty-aware prediction will be explored to improve robustness under distribution shifts and support deployment on the LuNiS lunar rover platform. Cross-platform evaluation across different rover morphologies and terrain conditions will further investigate the generalization capability of the proposed approach for future planetary exploration missions.


\section*{ACKNOWLEDGMENT}
This research was done in the SAMLER-KI project, funded by the German Federal Ministry for Economic Affairs and Energy (BMWE, grant number 50RA2203A). AI-assisted tools\cite{ChatGPT2026} were used for language editing.


\bibliographystyle{IEEEtranS}
\bibliography{SII}

@ARTICLE{Althoff,
  author={Althoff, Matthias and Dolan, John M.},
  journal={IEEE Transactions on Robotics}, 
  title={Online Verification of Automated Road Vehicles Using Reachability Analysis}, 
  year={2014},
  volume={30},
  number={4},
  pages={903-918},
  doi={10.1109/TRO.2014.2312453}}

@misc{ChatGPT2026,
  author       = {OpenAI},
  title        = {ChatGPT},
  year         = {2026},
  version      = {GPT-4}, 
  howpublished = {Large Language Model},
  url          = {https://chat.openai.com/}
}

@misc{fisac2018,
      title={A General Safety Framework for Learning-Based Control in Uncertain Robotic Systems}, 
      author={Jaime F. Fisac and Anayo K. Akametalu and Melanie N. Zeilinger and Shahab Kaynama and Jeremy Gillula and Claire J. Tomlin},
      year={2018},
      eprint={1705.01292},
      archivePrefix={arXiv},
      primaryClass={cs.RO},
      url={https://arxiv.org/abs/1705.01292}, 
}

@article{Fong,
author = {Fong, Terrence and Bualat, Maria and Deans, Matthew and Allan, Mark and Bouyssounouse, Xavier and Broxton, Michael and Edwards, Laurence and Elphic, Rick and Flückiger, Lorenzo and Frank, Jeremy and Keely, Leslie and Kobayashi, Leslie and Lee, Pascal and Lee, Susan and Lees, David and Pacis, E. and Park, Eric and Pedersen, Liam and Schreckenghost, D. and Utz, Hans},
year = {2008},
month = {09},
pages = {},
title = {Field Testing of Utility Robots for Lunar Surface Operations},
doi = {10.2514/6.2008-7886}
}

@article{Fuentes,
author = {Fuentes, Aishe and Kempf, Franziska and Kipfmüller, Martin and Bergmann, Tobias and Prieto, Miguel},
year = {2022},
month = {12},
pages = {44},
title = {Tip-Over Detection and Avoidance Algorithms as Stabilization Strategy for Small-Footprint and Lightweight Mobile Manipulators},
volume = {11},
journal = {Machines},
doi = {10.3390/machines11010044}
}

@article{Iwamoto,
author = {Iwamoto, Takumi and Shigematsu, Kosuke},
year = {2026},
month = {05},
pages = {1-6},
title = {DNN-Based Body Inclination Prediction for Tip-Over Prevention in Teleoperated Hydraulic Excavators},
volume = {7},
doi = {10.11532/jsceiiai.7.1_1},
journal = {Journal Name}
}

@article{Kang_2025,
   title={A real-time anomaly detection method for robots based on a flexible and sparse latent space},
   volume={158},
   ISSN={0952-1976},
   url={http://dx.doi.org/10.1016/j.engappai.2025.111310},
   DOI={10.1016/j.engappai.2025.111310},
   journal={Engineering Applications of Artificial Intelligence},
   publisher={Elsevier BV},
   author={Kang, Taewook and You, Bum-Jae and Park, Juyoun and Lee, Yisoo},
   year={2025},
 }

@misc{kendall2017,
      title={What Uncertainties Do We Need in Bayesian Deep Learning for Computer Vision?}, 
      author={Alex Kendall and Yarin Gal},
      year={2017},
      eprint={1703.04977},
      archivePrefix={arXiv},
      primaryClass={cs.CV},
      url={https://arxiv.org/abs/1703.04977}, 
}

@misc{malhotra2016lstmbasedencoderdecodermultisensoranomaly,
      title={LSTM-based Encoder-Decoder for Multi-sensor Anomaly Detection}, 
      author={Pankaj Malhotra and Anusha Ramakrishnan and Gaurangi Anand and Lovekesh Vig and Puneet Agarwal and Gautam Shroff},
      year={2016},
      eprint={1607.00148},
      archivePrefix={arXiv},
      primaryClass={cs.AI},
      url={https://arxiv.org/abs/1607.00148}, 
}

@inproceedings{Manderson,
author = {Manderson, Travis and Wapnick, Stefan and Meger, David and Dudek, Gregory},
year = {2020},
month = {05},
pages = {1263-1269},
title = {Learning to Drive Off Road on Smooth Terrain in Unstructured Environments Using an On-Board Camera and Sparse Aerial Images},
doi = {10.1109/ICRA40945.2020.9196879},
  booktitle = {Proceedings of the IEEE International Conference on Robotics and Automation (ICRA)}
}

@article{pub11033,
    author = {Wirkus, Malte and Arnold, Sascha
 and Berghöfer, Elmar
},
    title = {Online Reconfiguration of Distributed Robot Control Systems for Modular Robot Behavior Implementation},
    year = {2020},
    month = {12},
    volume = {100},
    number = {3},
    pages = {1283--1308},
    journal = {Journal of Intelligent \& Robotic Systems (JIRS)},
    publisher = {Springer Publishing}
}

@article{Mars2020,
author = {Farley, Kenneth and Williford, Kenneth and Stack, Kathryn and Chen, Allen and Torre, Manuel and Hand, Kevin and Goreva, Y. and Herd, Christopher and Hueso, Ricardo and Liu, Yang and Maki, Justin and Martinez, German M. and Moeller, Robert and Nelessen, Adam and Newman, Claire and Nunes, Daniel and Ponce, Adrian and Spanovich, Nicole and Wiens, Roger},
year = {2020},
month = {12},
pages = {},
title = {Mars 2020 Mission Overview},
volume = {216},
journal = {Space Science Reviews},
doi = {10.1007/s11214-020-00762-y}
}

@article{MCGHEE1968331,
title = {On the stability properties of quadruped creeping gaits},
journal = {Mathematical Biosciences},
volume = {3},
pages = {331-351},
year = {1968},
issn = {0025-5564},
doi = {https://doi.org/10.1016/0025-5564(68)90090-4},
url = {https://www.sciencedirect.com/science/article/pii/0025556468900904},
author = {R.B. McGhee and A.A. Frank}
}

@article{Pang_2021,
   title={Deep Learning for Anomaly Detection: A Review},
   volume={54},
   ISSN={1557-7341},
   url={http://dx.doi.org/10.1145/3439950},
   DOI={10.1145/3439950},
   number={2},
   journal={ACM Computing Surveys},
   publisher={Association for Computing Machinery (ACM)},
   author={Pang, Guansong and Shen, Chunhua and Cao, Longbing and Hengel, Anton Van Den},
   year={2021},
 }

@inproceedings{rauch2013comparison,
  title={Comparison of sensor-feedback prediction methods for robust behavior execution},
  author={Rauch, Christian and Bergh{\"o}fer, Elmar and K{\"o}hler, Tim and Kirchner, Frank},
  booktitle={Annual Conference on Artificial Intelligence},
  pages={200--211},
  year={2013},
  organization={Springer}
}

@article{Reviewonspacerobotics,
author = {Gao, Yang and Chien, Steve},
year = {2017},
month = {06},
pages = {eaan5074},
title = {Review on space robotics: Toward top-level science through space exploration},
volume = {2},
journal = {Science Robotics},
doi = {10.1126/scirobotics.aan5074}
}

@unknown{Rose,
author = {Rose, Nathaniel and Ahmed, Arif and Gutierrez-Cornejo, Emanuel and Maini, Parikshit},
year = {2025},
month = {07},
pages = {},
title = {Learning to Predict Mobile Robot Stability in Off-Road Environments},
doi = {10.48550/arXiv.2507.12731}
}

@article{Shaojie,
  author       = {Shaojie Bai and
                  J. Zico Kolter and
                  Vladlen Koltun},
  title        = {An Empirical Evaluation of Generic Convolutional and Recurrent Networks
                  for Sequence Modeling},
  journal      = {CoRR},
  volume       = {abs/1803.01271},
  year         = {2018},
  url          = {http://arxiv.org/abs/1803.01271},
  eprinttype   = {arXiv},
  eprint       = {1803.01271},
  bibsource    = {dblp computer science bibliography, https://dblp.org}
}

@Article{shete,
AUTHOR = {Shete, Siddhant and Domínguez, Raúl and Selvaraju, Ravisankar and Álvarez, Mariela De Lucas and Kirchner, Frank},
TITLE = {Prediction-Based Tip-Over Prevention for Planetary Exploration Rovers},
JOURNAL = {Engineering Proceedings},
VOLUME = {90},
YEAR = {2025},
NUMBER = {1},
ARTICLE-NUMBER = {44},
URL = {https://www.mdpi.com/2673-4591/90/1/44},
ISSN = {2673-4591}
}

@misc{shete_2026_21258126,
  author       = {Shete, Siddhant and
                  Domínguez, Raúl},
  title        = {IMU-Based Tip-Over Dataset for Space Exploration
                   Rover Dynamics and Stability Analysis
                  },
  month        = jul,
  year         = 2026,
  publisher    = {Zenodo},
  doi          = {10.5281/zenodo.21258126},
  url          = {https://doi.org/10.5281/zenodo.21258126},
}

@article{Wellhausen_2020,
   title={Safe Robot Navigation Via Multi-Modal Anomaly Detection},
   volume={5},
   ISSN={2377-3774},
   url={http://dx.doi.org/10.1109/LRA.2020.2967706},
   DOI={10.1109/lra.2020.2967706},
   number={2},
   journal={IEEE Robotics and Automation Letters},
   publisher={Institute of Electrical and Electronics Engineers (IEEE)},
   author={Wellhausen, Lorenz and Ranftl, Rene and Hutter, Marco},
   year={2020},
   month=Apr, pages={1326–1333} }

@misc{yu2020gradientsurgerymultitasklearning,
      title={Gradient Surgery for Multi-Task Learning}, 
      author={Tianhe Yu and Saurabh Kumar and Abhishek Gupta and Sergey Levine and Karol Hausman and Chelsea Finn},
      year={2020},
      eprint={2001.06782},
      archivePrefix={arXiv},
      primaryClass={cs.LG},
      url={https://arxiv.org/abs/2001.06782}, 
}

@article{ZMP,
author = {Vukobratovic, Miomir and Borovac, Branislav},
year = {2004},
month = {03},
pages = {157-173},
title = {Zero-Moment Point - Thirty Five Years of its Life.},
volume = {1},
journal = {I. J. Humanoid Robotics},
doi = {10.1142/S0219843604000083}
}


\end{document}